\documentclass{article}

\usepackage[final]{neurips_2024}

\usepackage[utf8]{inputenc}
\usepackage[T1]{fontenc}

\usepackage{amsmath}
\usepackage{amsfonts}

\usepackage{hyperref}
\usepackage{url}
\usepackage{tabularx} 
\usepackage{booktabs}
\usepackage{nicefrac}
\usepackage{microtype}
\usepackage{xcolor}
\usepackage{graphicx}
\usepackage{subcaption}

\usepackage{booktabs}
\usepackage{listings}
\usepackage{xcolor}
\usepackage{float}        
\usepackage{placeins}     
\usepackage{dblfloatfix}  

\title{Observer, Not Player: Simulating Theory of Mind in LLMs through Game Observation}

\author{
Jerry Wang, Ting Yiu Liu\\
Department of Management Information Systems,\\ National ChengChi University\\
\texttt{111306078@nccu.edu.tw, 113356048@nccu.edu.tw}
}

\makeatletter
\renewcommand{\maketitle}{\@maketitle}
\makeatother

\begin{document}
\maketitle
\begin{abstract}
We present an interactive framework for evaluating whether large language models (LLMs) exhibit genuine ``understanding'' in a simple yet strategic environment. As a running example, we focus on Rock--Paper--Scissors (RPS), which, despite its apparent simplicity, requires sequential reasoning, adaptation, and strategy recognition. Our system positions the LLM as an \emph{Observer} whose task is to identify which strategies are being played and to articulate the reasoning behind this judgment. The purpose is not to test knowledge of Rock--Paper--Scissors itself, but to probe whether the model can exhibit mind-like reasoning about sequential behavior.
To support systematic evaluation, we provide a benchmark consisting of both static strategies and lightweight dynamic strategies specified by well-prompted rules. We quantify alignment between the Observer’s predictions and the ground-truth distributions induced by actual strategy pairs using three complementary signals: Cross-Entropy, Brier score, and Expected Value (EV) discrepancy. These metrics are further integrated into a unified score, the \emph{Union Loss}, which balances calibration, sensitivity, and payoff alignment. Together with a Strategy Identification Rate (SIR) metric, our framework captures not only predictive accuracy but also whether the model can stably identify the latent strategies in play.
The demo emphasizes \textbf{interactivity, transparency, and reproducibility}. Users can adjust LLM distributions in real time, visualize losses as they evolve, and directly inspect reasoning snippets to identify where and why failures occur. In doing so, our system provides a practical and interpretable proxy for mind-like inference in sequential games, offering insights into both the strengths and limitations of current LLM reasoning.

\end{abstract}

\section{Motivation and Goal}
\label{sec:intro}

Evaluating what a large language model (LLM) has truly learned in an interactive setting is challenging, because real-world mechanisms are complex and seldom come with explicit explanations. Games provide a useful test bed because their rules are transparent and can be regarded as ground truth. Also, the environment is controllable, and interactions can be reproduced through prompting. For these reasons, games have long served as benchmarks for studying AI behavior.

Recent studies show that LLMs can play games, craft strategies, and adapt to varied environments, spanning board games, repeated social dilemmas, simulations, and program syntheses~\cite{bhatia2024playing,mcaleese2024repeated,jun2024can,liu2024lmgamebench,wang2024adaptive,prystawski2024code,lei2024idge}. In most of these works, models act as agents that interact with opponents. Because evaluation often relies only on overall win rate, it is hard to tell whether a model understands the game or merely adapts to observed losses. Win rate compresses pattern recognition, planning, opponent modeling, and luck into a single number, so a high score does not guarantee that the model captures the underlying outcome distribution against each opponent strategy. When the opponent or the initial conditions change, metrics based on win rate become even less reliable. This issue connects to the broader debate on whether LLMs exhibit  \emph{Theory of Mind (ToM)}–like reasoning or simply exploit dataset artifacts and surface cues~\cite{Kosinski_2024,ullman2023largelanguagemodelsfail}.

To obtain a clearer picture of the model’s capabilities, we let the LLM act solely as an \emph{Observer}. For each matchup, the model is tasked with predicting the probability distribution over candidate strategies in \emph{Rock--Paper--Scissors (RPS)}, and we compare its forecast against the ground-truth distribution generated by the game engine. To provide the LLM with sufficient information for reasoning, our framework encapsulates the ground-truth rationale, current trajectory round, and trajectory history within a comprehensive \emph{chain-of-thought (CoT)} prompt. This unified context leverages recent advances in prompting techniques, which have significantly enhanced LLM reasoning abilities. Specifically, chain-of-thought prompting~\cite{wei2022chainofthought} introduces intermediate reasoning steps into few-shot exemplars, enabling LLMs to solve complex tasks with greater accuracy. Recent paradigms such as Coconut~\cite{hao2024coconut} extend this by supporting reasoning in continuous latent spaces, while Zero-Shot CoT~\cite{kojima2022zeroshotcot} demonstrates that even minimal prompt engineering can elicit strong step-by-step reasoning in large models. For practical deployment and real-time interpretability, we adopt standard zero-shot CoT prompting. This allows each round’s prediction and reasoning process to be transparently displayed to users, rather than relying on more complex or opaque paradigms such as Coconut.
For evaluation, we operationalize “game understanding” as distributional alignment between the model's guesses and the ground-truth outcome distribution.We quantify this alignment using three proper scoring metrics: \emph{Cross-Entropy (CE)}~\cite{kullback1951information,shannon1948mathematical,goodfellow2016deep}, the Brier score~\cite{brier1950verification}, and an \emph{Expected-Value discrepancy (EV)}~\cite{gneiting2007strictly,dawid1984statistical}.
ceevTo ensure comparability, we apply fixed-bound normalization to the EV term, no further adjustment to the Brier score since it already lies in $[0,1]$, and grid-wise min–max normalization to the cross-entropy term, and then average the three metrics to define the \textbf{Union Metric}. A model that genuinely understands the matchup should achieve low values on every component, resulting in a low Union Loss.

We selected Rock–Paper–Scissors for its combination of simplicity and strategic depth. Its transparent rules and non‐transitive payoff structure allow us to populate our strategy pool with a diverse set of opponents including static strategies with pure and biased to represent  dynamic uniform strategies agents without a fixed plan but sampling with bias, and human‐inspired policies such as “win last,” “lose last,” and “copy last”. Moreover, any two strategies admit a closed-form outcome distribution and  can be analyzed efficiently via trajectory approximations. The action space also makes RPS ideal for supporting on-laptop demos aligned with a demo-track’s goals.

\begin{figure}[t]
  \centering
  \begin{subfigure}[b]{0.45\linewidth}
    \includegraphics[width=\linewidth]{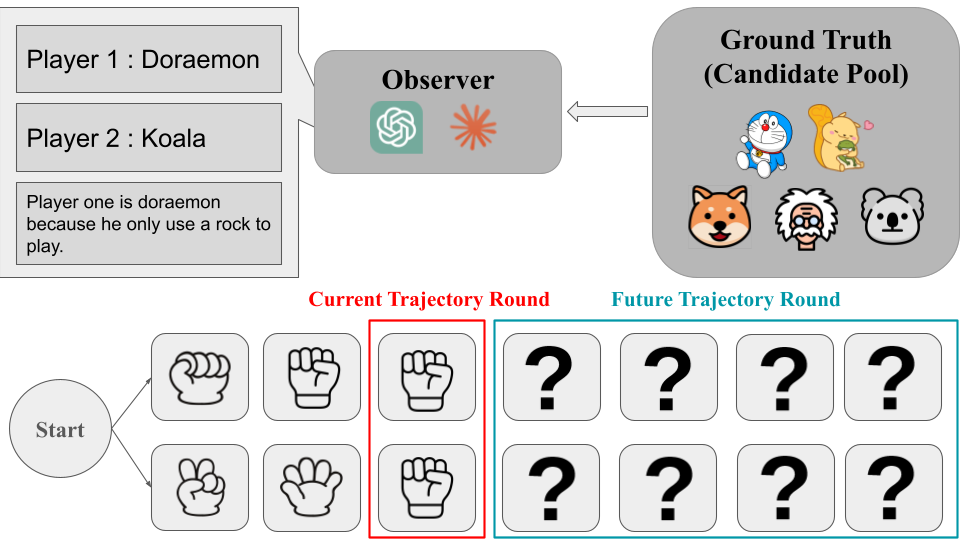}
    \caption{Early-round forecasts by the LLM Observer}
    \label{fig:early}
  \end{subfigure}
  \hfill
  \begin{subfigure}[b]{0.45\linewidth}
    \includegraphics[width=\linewidth]{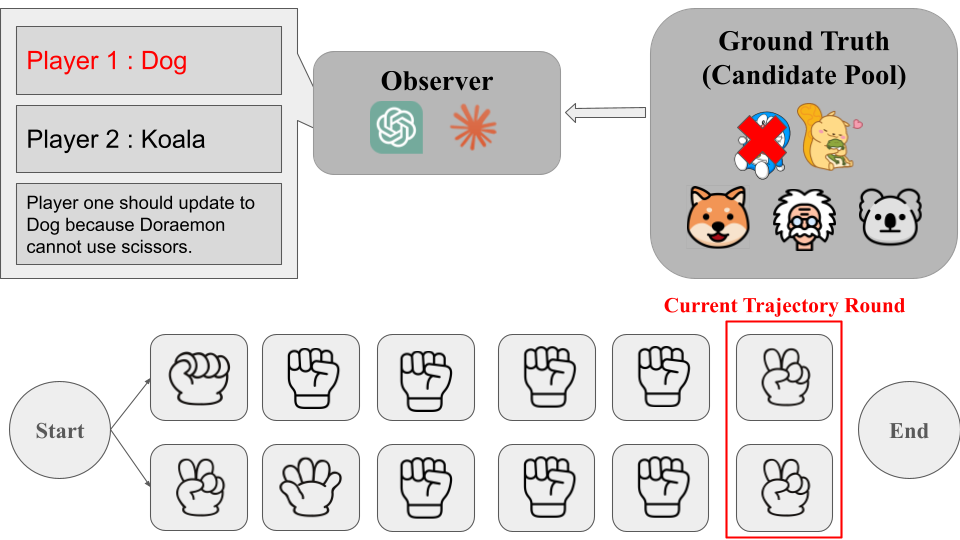}
    \caption{Late-round update of previous strategy}
    \label{fig:late}
  \end{subfigure}
  \caption{Evolution of the LLM Observer’s probabilistic forecasts over the course of the match.}
  \label{fig:observer_evolution}
\end{figure}

Figure~\ref{fig:early} and Figure~\ref{fig:late} illustrate how LLMs serve as observers in the gaming environment. In Figure~\ref{fig:early}, at the start of a match, the observer issues an initial forecast after several rounds. This guess may assign different thought from possibility from both players in this example or even gives strange thoughts in other cases, reflecting a biased hypothesis based on sparse evidence on current trajectory rounds. As the game unfolds (Figure~\ref{fig:late}), guesses are continuously compared with the true outcome distribution.forecast By the late rounds, both player provides more gaming history so the previous hypotheses are discarded. Through this dynamic process, we observe how the LLM updates its beliefs and converges on the most likely outcome. This mechanism mirrors how humans form early assumptions and revise them when new information arrives.

\paragraph{Goal of the demo.}
Our goal is not to chase state-of-the-art benchmarks but to deliver an interactive, transparent and reproducible platform for probing distributional understanding in a simple strategic environment. We provide a real-time web interface that lets users configure static and dynamic opponents, then view the Observer’s probability forecasts alongside proper-scoring-rule losses. Live visual analytics highlight where and why the model’s beliefs diverge from the true outcome distribution. The lightweight implementation promotes community reuse and reproducibility. The design also supports ablation studies—such as varying prompts or temperature settings—and can be extended with methods like change-point detection for strategy switches~\cite{adams2007bocpd}. Ultimately, the demo demonstrates how an LLM forms, updates and refines its beliefs when interacting with a game-like environment, which also have potential to become benchmarks for theory of mind research.

Theory-of-Mind ()forecast
\section{Related Work}

\subsection{LLM ToM Evaluation}
A growing body of work probes whether large language models exhibit theory-of-mind-like competencies~\cite{strachan2024nature}. Beyond single-task probes, ToMBench systematizes ToM evaluation into 8 task families and 31 abilities to mitigate leakage and subjective grading~\cite{chen2024tombenchbenchmarkingtheorymind}. In addition, recent work benchmarks 11 models against children 7--10 years old on richer ToM inventories~\cite{van-duijn-etal-2023-theory}. Recent position papers argue that many benchmarks capture \emph{literal} ToM (predicting others) rather than \emph{functional} ToM (adapting to new partners), urging interactive settings that test adaptation, calibration, and belief updating~\cite{riemer2025positiontheorymindbenchmarks}. Our setup complements this line by treating “understanding” as alignment to \emph{outcome distributions} rather than pass/fail answers, and by making uncertainty explicit through proper scoring rules.

\subsection{Interactive Visualization Platforms for Games and Agents}
Interactive, code-light tools have long supported transparent debugging and diagnosis. TensorBoard popularized real-time tracking and embedding projections for deep models~\cite{abadi2016tensorflowlargescalemachinelearning}, while the What-If Tool introduced point-and-click counterfactuals and per-example analysis without code changes~\cite{Wexler_2019}. For RL and games, OpenAI Gym standardized interfaces and built-in monitoring/video capture~\cite{brockman2016openaigym}; PettingZoo unified multi-agent environments under an AEC API~\cite{terry2021pettingzoogymmultiagentreinforcement}; OpenSpiel offered a research platform for many games with evaluation utilities~\cite{lanctot2020openspielframeworkreinforcementlearning}; RLCard added card-game suites and visual tools~\cite{zha2020rlcardtoolkitreinforcementlearning}; Unity ML-Agents provided a general 3D simulation stack with built-in viewers~\cite{juliani2020unitygeneralplatformintelligent}. Visualization systems tailored to RL (e.g., Vizarel/DRLViz) further target policy rollouts, memory inspection, and failure analysis~\cite{deshpande2020interactivevisualizationdebuggingrl}. Our dashboard extends this tradition to \emph{distributional} diagnostics in RPS.

\subsection{Game-Based Benchmarks for Agent Evaluation}
Simple strategic games offer controllable yet revealing tests of planning, opponent modeling, and adaptation. \emph{SmartPlay} curates multiple games (incl.\ RPS) to isolate nine agent capabilities and multi-turn generalization~\cite{wu2024smartplaybenchmarkllmsintelligent}. \emph{GameBench} targets LLM strategic reasoning across nine game environments~\cite{costarelli2024gamebenchevaluatingstrategicreasoning}, and \emph{AgentBench} evaluates LLMs-as-agents across diverse interactive tasks~\cite{liu2023agentbenchevaluatingllmsagents}. Cooperative and imperfect-information games provide incisive stress tests: the Hanabi Challenge foregrounds belief inference and partner modeling~\cite{Bard_2020}; Overcooked-AI probes human–AI coordination under time pressure~\cite{carroll2020utilitylearninghumanshumanai}. Text-based and gridworld platforms add language and compositionality: TextWorld for language-conditioned RL in generated games~\cite{côté2019textworldlearningenvironmenttextbased}, BabyAI for grounded-language curricula~\cite{chevalierboisvert2019babyaiplatformstudysample}, and the NetHack Learning Environment for long-horizon, procedural play with rich state/action spaces~\cite{küttler2020nethacklearningenvironment}. Recent LLM-focused game suites further emphasize turn-by-turn logging and leaderboarded play on grid-based games such as Tic-Tac-Toe, Connect Four, and Gomoku~\cite{topsakal2024evaluatinglargelanguagemodels}. Relative to these, our demo centers not on win rate, but on \emph{probabilistic guessing} of outcomes against static/dynamic opponents, exposing where models’ beliefs diverge from ground truth.


\section{System Overview}
\label{sec:system}
\begin{figure}[!htbp]
  \centering
  \includegraphics[width=0.9\linewidth]{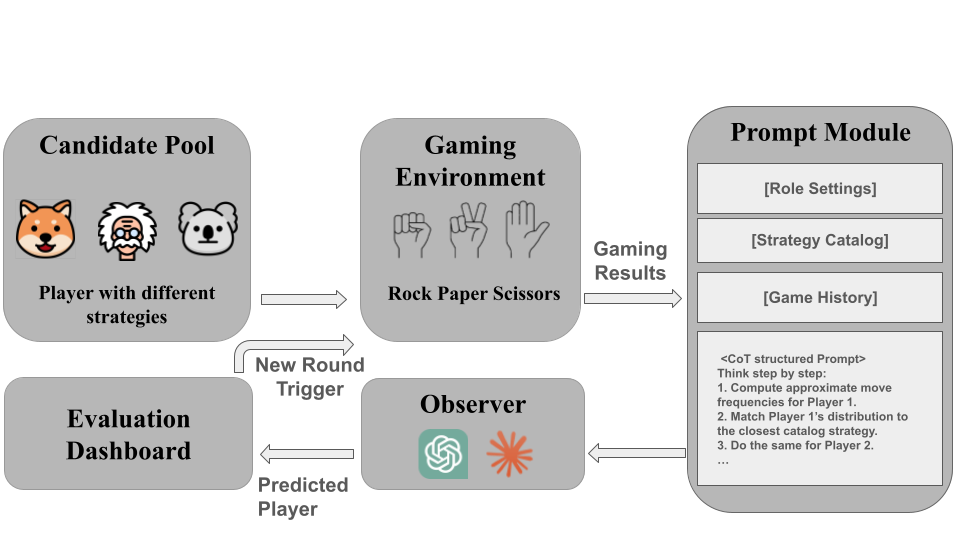}
  \caption{\textbf{System pipeline.}  
          Solid arrows indicate the data flow from user-selected strategies in the Candidate Pool, through the RPS engine and Prompt Module, to the LLM Observer and the real-time Evaluation Dashboard.}
  \label{fig:system}
\end{figure}

\paragraph{Workflow and Steady–State Solver.}
Figure~\ref{fig:system} shows the real‐time pipeline. The \textbf{Candidate Pool} (App.~\ref{app:player-table}) is organized into three distinct classes to capture different behavioral patterns. First, we include \emph{Human (reactive) policies} (X, Y, Z) that update as a function of the opponent’s distribution or the previous outcome—namely “win‐last,” “lose‐last,” and “copy‐last”—to model simple, memory‐dependent behaviors. Next, we consider \emph{static strategies} (A, B, C), which are pure Rock, Paper, or Scissors distributions serving as analytically tractable baselines. Finally, the remaining entries (D–P) are \emph{biased dynamic mixtures} with fixed probability vectors that interpolate or bias the pure actions. Only the adaptive strategies require an update map \(g_k:\Delta^3\!\to\!\Delta^3\).  

For the steady‐state approximation, let \(s_i^{(t)}\!\in\!\Delta^3\) denote player~\(i\)’s mixed strategy at solver iteration~\(t\), initialized at \(s_i^{(0)}=(1/3,1/3,1/3)\). If player~\(i\) is adaptive (X, Y, Z), we apply a damped fixed‐point iteration
\begin{equation}
  s_i^{(t+1)} \;=\; \alpha\,g_{k_i}\!\bigl(s_{-i}^{(t)}\bigr) \;+\; (1-\alpha)\,s_i^{(t)},
  \qquad \alpha\in(0,1),
  \label{eq:damped-update}
\end{equation}
while static or biased mixture strategies keep \(s_i^{(t+1)}=s_i^{(t)}\).  

In the fully co‐adaptive regime, we simulate two adaptive agents through the coupled updates
\begin{align}
  s_1^{(t+1)} &= \alpha\,g_{k_1}\!\bigl(s_2^{(t)}\bigr) \;+\; (1-\alpha)\,s_1^{(t)}, 
  \label{eq:update1-dd} \\
  s_2^{(t+1)} &= \alpha\,g_{k_2}\!\bigl(s_1^{(t)}\bigr) \;+\; (1-\alpha)\,s_2^{(t)}.
  \label{eq:update2-dd}
\end{align}
Iterations stop when 
\begin{equation}
  \|s_i^{(t+1)}-s_i^{(t)}\|_1 < 10^{-4}, 
  \label{eq:converge-crit}
\end{equation}
for all adaptive players or when a preset cap is reached. We empirically observe convergence for the tested settings (formal guarantees are out of scope). Let \((s_1^*,s_2^*)\) be the resulting steady state; the outcome probabilities 
\(\Pr[\mathrm{win}], \Pr[\mathrm{draw}], \Pr[\mathrm{loss}]\) are then computed under this stationary‐mixing approximation and fed to the Observer’s loss.

The current game state and cumulative history are provided to the \textbf{Prompt Module}, which constructs a four-part chain-of-thought (CoT) prompt consisting of: (1) candidate information, (2) role specification, (3) previous game trajectory results, and (4) a request for the model to predict the next player action with explicit reasoning. The CoT prompt enables the language model to articulate its decision-making process, thereby providing transparency into \emph{why} a particular choice was made. Furthermore, by encapsulating the full game context within the prompt, our approach ensures that evaluation measures the model’s reasoning in context, rather than relying on any latent background knowledge. Specific prompt we used for experiment can be found in Appendix ~\ref{app:prompt-example}.

The completed prompt is routed to the \textbf{Observer}, an LLM selected at run time, executing the task and generating guesses for both players, with all metrics streamed to the \textbf{Evaluation Dashboard}, a lightweight web interface that renders updated result
 for current model loss, enabling real-time prompt
ablations or temperature sweeps.

\section{Evaluation Mechanisms}
\label{sec:metrics}

To judge whether an LLM \emph{truly understands} the game---rather than merely fitting surface patterns—we evaluate its predictions at both the distributional and payoff levels via several metrics below. For each matchup, let $p^\star = (p^\star_{\text{win}}, p^\star_{\text{draw}}, p^\star_{\text{loss}})$ denote the ground-truth outcome distribution, and let $\hat{p}$ denote the probability distribution over win, draw, and loss outcomes as predicted by the LLM Observer after modeling the interaction between the two players’ strategies forecast.

We report three complementary metrics. First, cross-entropy (CE )measures the information-theoretic ``surprise'' when the true distribution is encoded by the model's prediction, rewarding forecast that place high probability on the correct outcomes:
\begin{equation}\label{eq:ce}
  \mathrm{CE}(p^\star,\hat{p}) = -\sum_{c \in \{\mathrm{win,draw,loss}\}} p^\star_c\,\log(\hat{p}_c + \varepsilon),
\end{equation}
where the constant $\varepsilon$ prevents numerical underflow. Note that CE is not guaranteed to be zero even when the predicted distribution perfectly matches the true distribution: unless the ground truth is deterministic (e.g., one outcome with probability $1$), CE will equal the entropy of the true distribution, which is strictly positive.

Second, the \emph{Brier score} penalizes both mis-ranking and mis-calibration of probabilities, serving as a calibration-sensitive proper scoring rule:
\begin{equation}\label{eq:brier}
  \mathrm{Brier}(p^\star,\hat{p}) = \sum_{c \in \{\mathrm{win,draw,loss}\}} (\hat{p}_c - p^\star_c)^2.
\end{equation}
Whereas CE and Brier together capture distributional alignment (i.e., whether the model predicts the full outcome distribution correctly and with proper confidence), they may still overlook systematic errors in payoff assessment.

To address this, we further compute the \emph{expected-value discrepancy} (EVLoss), which directly checks whether the model over- or underestimates the net advantage implied by its forecast:
\[
  \mathrm{EV}(p) = \frac{p_{\text{win}} - p_{\text{loss}}}{100},
\]
\begin{equation}\label{eq:evloss}
  \mathrm{EVLoss}(p^\star,\hat{p}) = \left(\mathrm{EV}(p^\star) - \mathrm{EV}(\hat{p})\right)^2.
\end{equation}
The theoretical range of $\mathrm{EV}(p)$ is $[-1,1]$, since wins and losses can differ by at most 100 out of 100 matches. Therefore, the discrepancy $\mathrm{EV}(p^\star) - \mathrm{EV}(\hat{p})$ lies in $[-2,2]$, and $\mathrm{EVLoss} \in [0,4]$. The lower bound $0$ occurs when the predicted and true expected values coincide, while the upper bound $4$ occurs when they are maximally opposed (e.g., $\mathrm{EV}(p^\star)=1$ and $\mathrm{EV}(\hat{p})=-1$).

Finally, we combine the three components into a single metric, \emph{Union Loss}, by averaging normalized scores:
\begin{equation}\label{eq:union}
  \mathrm{Union}(p^\star,\hat{p}) = \frac{1}{3}\left[\mathrm{CE}_{\text{norm}}(p^\star,\hat{p}) + \mathrm{Brier}_{\text{norm}}(p^\star,\hat{p}) + \mathrm{EVLoss}_{\text{norm}}(p^\star,\hat{p})\right].
\end{equation}
For comparability, we normalize each component differently according to its natural range:  
(i) $\mathrm{EVLoss}$ is divided by $4.0$, its theoretical maximum;  
(ii) $\mathrm{Brier}$ already lies in $[0,1]$ and is left unchanged;  
(iii) $\mathrm{CE}$ has no fixed bound and is normalized via grid-wise min--max scaling across all matchups:
\[
\mathrm{CE}_{\text{norm}}(x) = 
\begin{cases}
\frac{x - \min(\mathrm{CE})}{\max(\mathrm{CE}) - \min(\mathrm{CE})}, & \max(\mathrm{CE}) \neq \min(\mathrm{CE}), \\
0.5, & \max(\mathrm{CE}) = \min(\mathrm{CE}).
\end{cases}
\]

\begin{figure}[!htbp]
  \centering
  \includegraphics[width=\linewidth]{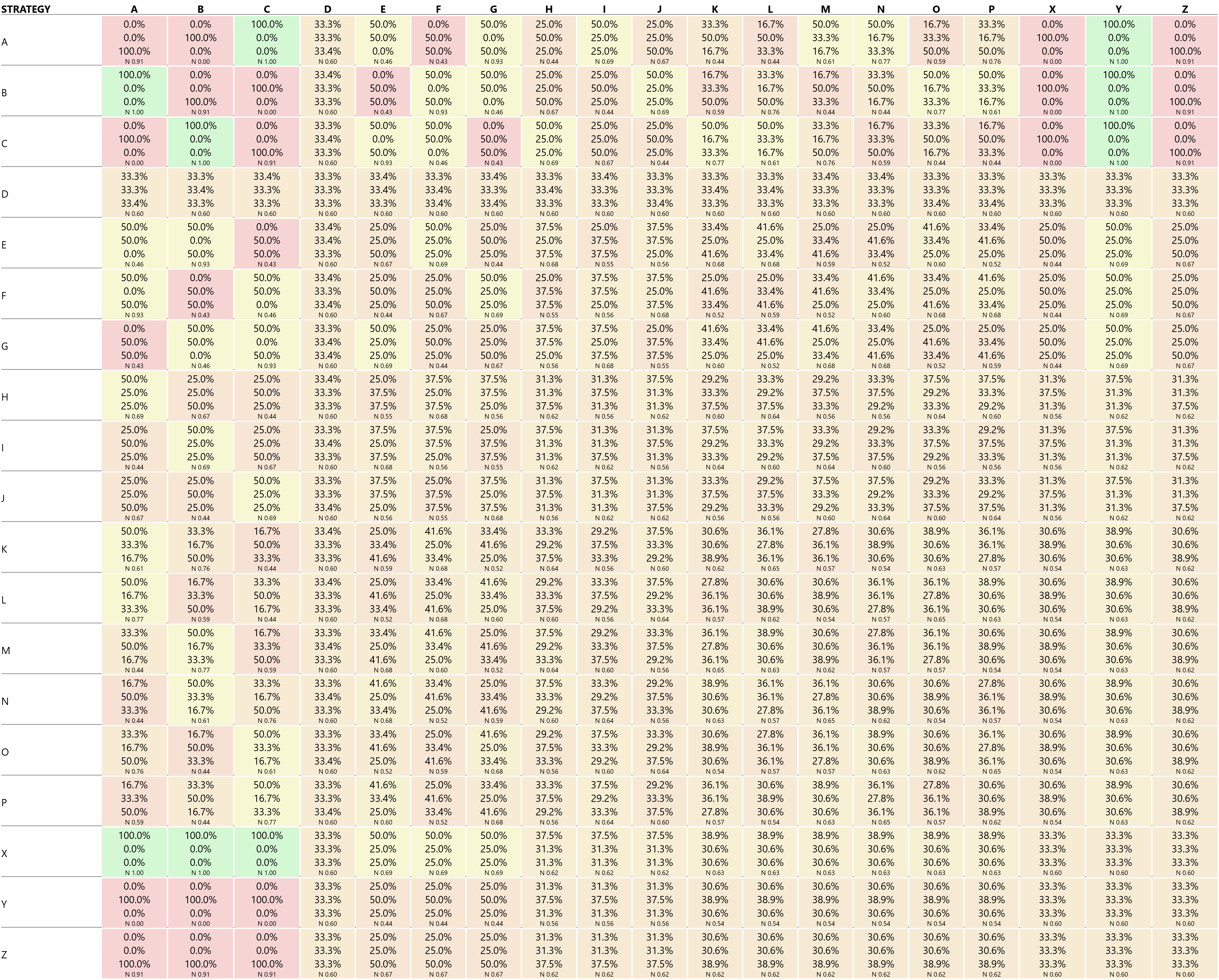}
  \caption{Heatmap visualization of loss values across strategy matchup. 
  Green cells indicate predictions that closely match the actual outcome distribution, yellow to orange denote moderate deviations, and red highlights severe mismatches.}
  \label{fig:heatmap}
\end{figure}

These metrics jointly evaluate distinct aspects of model reasoning, each corresponding to an essential component of theory of mind. Specifically, cross-entropy and Brier score probe the \emph{behavioral layer}: does the model accurately predict the full outcome distribution and express appropriately calibrated confidence? In contrast, EV Loss targets the \emph{utility layer} to show whether the model infers both the direction and magnitude of the expected payoff. A low Union Loss thus indicates that the model aligns with both action prediction (cognition) and utility inference—the dual capacities fundamental to robust theory-of-mind–style game understanding. 
\\
To provide better intuition of how the framework operates, we visualize the outcome space as a heatmap. Figure~\ref{fig:heatmap} illustrates how the loss behaves when the LLM makes different predictions against the ground‐truth strategy pair. Each block shows the corresponding loss relative to the actual outcome, with color intensity encoding the degree of deviation. As an illustrative example, we highlight a combat scenario between strategy~A and strategy~B, where the heatmap clearly demonstrates that mismatched guesses incur larger penalties, whereas accurate predictions correspond to lower loss. This visualization offers an intuitive way for human readers to assess the quality of the Observer Module’s reasoning across the candidate pool and serves as a convenient diagnostic tool.

\section{Experimental Setting}
\label{sec:exp-setting}

Our study is guided by two central research questions (RQs). 
First, \emph{do LLM reasoning traces reliably yield correct sequential strategy inferences?} 
Second, \emph{how do our proposed metrics relate to mind‐like inference?} 
To address these questions, we evaluate each agent under the three interaction patterns introduced in Sec.~\ref{sec:system}.

\paragraph{Agents.}
We test three recent instruction‐tuned models \textsc{GPT-4o-mini}, \textsc{o3}, and \emph{\textsc{Claude 3.7 Sonnet} (Claude)}~\cite{openai2024gpt4ocard}.
These models were selected to provide coverage across (i) different providers (OpenAI vs. Anthropic),
(ii) different cost–latency tiers (lightweight ``mini'' vs. reasoning-optimized ``o3''), and
(iii) different design philosophies in alignment and reasoning.
This diversity allows us to examine whether distributional understanding is consistent across
architectures, scales, and training paradigms, rather than being confined to a single family of models.

\paragraph{Ground truth.}
We provided Table 3 in Appendix ~\ref{app:player-table} as the reference set of ground-truth strategies for all player roles. 
Each role corresponds to a predefined distribution over rock, paper, and scissors moves, ranging from pure strategies 
to mixed or biased ones, which serve as the canonical answer space against which the LLMs generate their guesses.

\paragraph{LLM prompting.}
We fix the chain-of-thought template in App.~\ref{app:prompt-example},
append the current round transcript, and query the model at temperature~0.2
(\textsc{Top-p}=2.0). We fix the chain-of-thought template in App.~\ref{app:prompt-example}, 
append the current round transcript, and query the model at temperature~0.2 
(\textsc{Top-p}=0.7). The model outputs its predictions for two players (\texttt{guess\_s1} and \texttt{guess\_s2}) together with a confidence score. In addition, every 20 rounds the model provides a brief reasoning summary, consisting of 3--5 phrases separated by semicolons, enabling us to monitor how its decision process evolves over time. We further introduce a warm-up phase by preloading 10 rounds of match data before the model begins making predictions. During inference, the model may reference at most the 50 most recent rounds as history (history limit). While models may “think” arbitrarily long, each response must conclude with the required output.

\paragraph{Match-up configurations and variables.}
We design three representative match-up regimes to expose models to distinct strategic dynamics: 

(1) \textbf{Dynamic vs.~Static (H vs.~C)}: $H$ is a rock-biased player with distribution $\{ \text{rock}:0.5, \text{paper}:0.25, \text{scissors}:0.25 \}$, while $C$ is a pure paper player with distribution $\{ \text{rock}:0, \text{paper}:1, \text{scissors}:0 \}$. 

(2) \textbf{Dynamic vs.~Dynamic (N vs.~G)}: $N$ favors paper primarily and scissors secondarily $\{ \text{rock}:0.167, \text{paper}:0.5, \text{scissors}:0.333 \}$, while $G$ mixes paper and scissors $\{ \text{rock}:0, \text{paper}:0.5, \text{scissors}:0.5 \}$.

(3) \textbf{Dynamic vs.~Psychological (D vs.~Y)}: $D$ plays paper-primary, scissors-secondary $\{ \text{rock}:0.167, \text{paper}:0.5, \text{scissors}:0.333 \}$, whereas $Y$ is adaptive, choosing the counter-move to the opponent’s most recent bias.

We treat the choice of \emph{model} as the independent variable, with three levels \{GPT-4o-mini, GPT-o3, Claude-3.7\}, and the match-up regime as the manipulated condition of true strategies, also with three levels \{H vs C, N vs G, D vs Y\}, yielding $3 \times 3 = 9$ total experimental conditions. All other parameters are controlled for comparability: the number of rounds is fixed at 200; a warm-up of 10 rounds allows the model to observe before issuing guesses; the history limit is set to 50 as the maximum number of past rounds the model may reference; and a reasoning interval of 20 requires the model to emit a brief reasoning summary every 20 rounds.\(\texttt{rounds}=\)\(\texttt{warmup\_rounds}=\)forecastwindow\(\texttt{history\_limit}=\)\(\texttt{reasoning\_interval}=\)

\paragraph{Evaluation metrics.}
For each pair of guesses $(p^\star, \hat{p})$, we evaluate performance using the metrics defined in Sec.~\ref{sec:metrics}: (1) cross-entropy, (2) Brier score, (3) EV discrepancy, and (4) their aggregate Union Loss.
Also, we consider an (5)\emph{explicit-commitment} metric (SIR) that requires correct identification of \emph{both} players’ strategies.forecastceMetric We report per-regime means $\pm$ standard error, and conduct one-sided permutation tests (10,000 resamples) against the human baseline.

\section{Demo Scenarios and Formative Findings}

\subsection{Evaluation of LLM Reasoning Traces for Sequential Strategy Inference (RQ1)}
\begin{figure}
    \centering
    \includegraphics[width=1.0\linewidth]{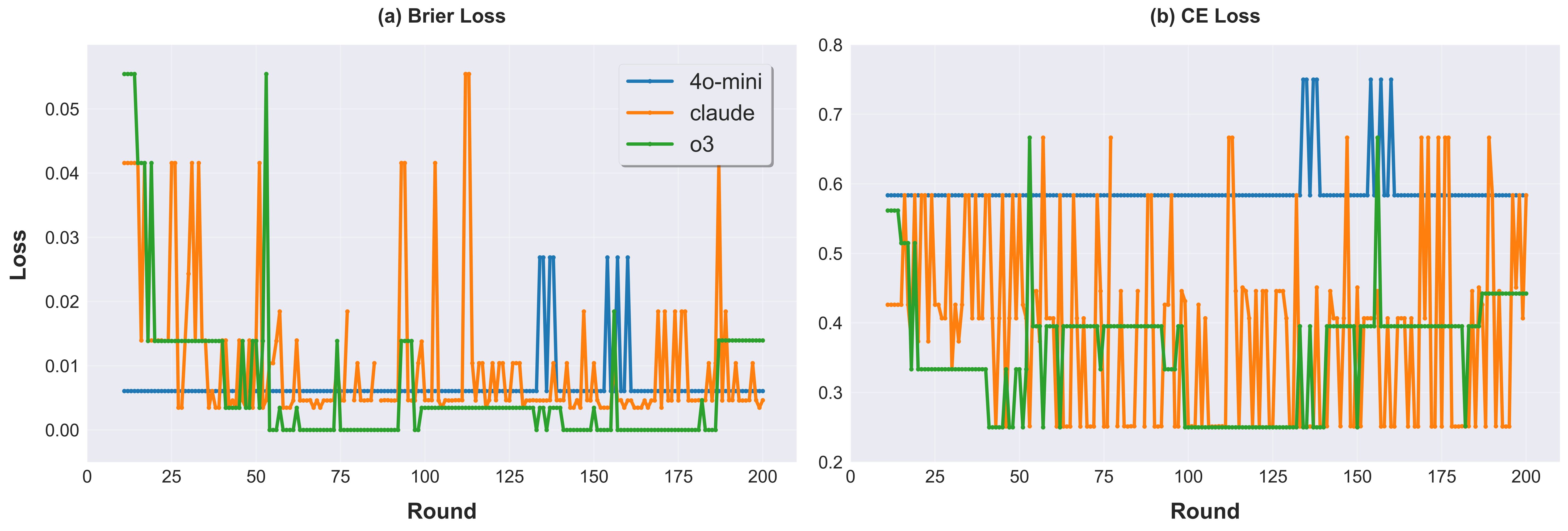}
    \caption{Brier loss and normalized Cross-Entropy loss for the N vs. G matchup}
    \label{fig:loss_all}
\end{figure}

Reasoning traces are often taken as evidence of “understanding,” yet it remains unclear whether such narratives \textit{reliably} lead to correct answers in sequential settings. To address this gap, we operationalize reliability through three complementary lenses: loss-based signals, an explicit-commitment metric (SIR), and model-provided reasoning. Together, these perspectives allow us to assess whether reasoning traces consistently yield correct sequential inferences. Agreement across these lenses would indicate convergent validity for our evaluation and, importantly, capture competencies that are necessary but not sufficient for mind-like inference.
\begin{figure}
    \centering
    \includegraphics[width=1.0\linewidth]{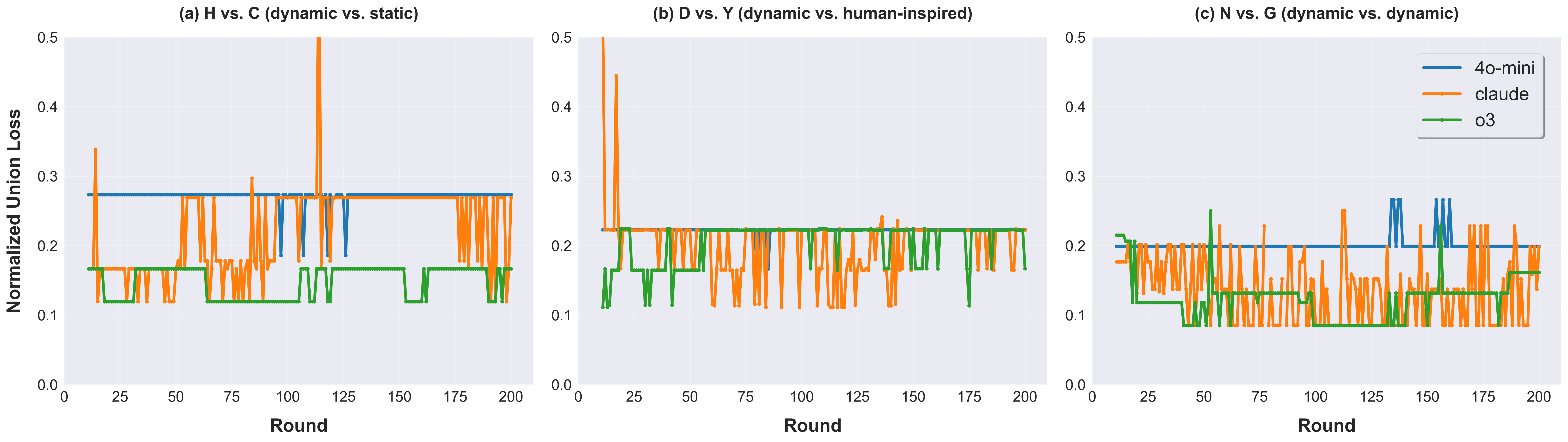}
    \caption{Comparison of Union Loss across models in different settings.}
    \label{fig:union_loss_all}
\end{figure}
Figure~\ref{fig:union_loss_all} and Figure~\ref{fig:loss_all} reveal clear round-wise differences and convergence patterns across the three models. Among these models, \textit{o3} exhibits a characteristic profile with a brief initial transient, a rapid decline to a low level, and a stable plateau with only small fluctuations. This trend is evident in the N–G matchup, where the Brier loss approaches zero after roughly 40–60 rounds and remains near that value, while the normalized cross-entropy decreases monotonically between rounds 50–100 and then stays low with only minor excursions. \textit{Claude} shows a mid-range yet high-variance profile. Although its losses tend to decline in the middle-to-late segments, the curves contain frequent spikes of varying magnitude, indicating high sensitivity to local histories and limited stability. In contrast, \textit{GPT-4o-mini} remains almost invariant across rounds. In the N–G matchup, the Brier loss stays as a nearly flat, low-amplitude baseline, whereas the normalized cross-entropy remains at a relatively high constant level with a pronounced spike only near rounds 140–160. This combination of a low and steady Brier loss together with a high and flat cross-entropy suggests that its predictive distribution is barely updated. The model behaves as if it emits a fixed, smoothed prior, thereby allocating insufficient probability mass to the true class and being penalized under log loss.
Taken together, these results provide a direct answer to \textbf{RQ1}. 
We find that only \textit{o3} demonstrates reliable sequential strategy inference, as indicated by rapid convergence and stability across both loss metrics. \textit{Claude} shows partial but unstable reliability due to high variance, while \textit{GPT-4o-mini} fails to adapt its predictions and therefore does not yield reliable inferences.

\subsection{Relation of Our Metrics to Mind-like Inference (RQ2)}
To assess whether our metrics capture \emph{ToM-relevant} competencies (rather than ToM itself), 
we triangulate the loss-based trajectories with a textual commitment check---the Strategy Identification Rate (SIR)---and qualitative analysis of the models’ own reasoning ~\ref{tab:rep_snippets},  see Appendix ~\ref{app:reasoning-data} for full tables. 

Across \autoref{fig:union_loss_all}, \autoref{tab:strategy_id} (SIR), and representative rationale excerpts (App.~\ref{app:reasoning-data}), we observe a consistent model ordering and round-wise dynamics.

\paragraph{o3.}
o3's rationales follow a systematic pattern: it first quantifies round-wise frequencies, then maps them to the strategy catalog by citing proximity (e.g., smallest distance to $G$ for \emph{Paper+Scissors}), and explicitly \emph{rules out} dynamic rules with concrete evidence (e.g., “zero Rock precludes X/Y/Z”). 
This disciplined style yields \emph{stable static attributions} as the window expands, high SIR in the Static--Dynamic and Dynamic--Dynamic settings (see \autoref{tab:strategy_id}), and \emph{low, comparatively smooth} Union Loss in \autoref{fig:union_loss_all}(a,b). 
Even when facing Random vs.\ Counter-Last (D vs.\ Y), where success requires tracking \emph{opponent-conditioned, lag-1 contingencies}, o3 remains comparatively strongest, although all models show elevated loss and near-zero SIR (\autoref{fig:union_loss_all}c; \autoref{tab:strategy_id}).
\begin{table}[ht]
  \centering
  \caption{\textbf{Strategy Identification Rate (\%).}
  Values denote the percentage of rounds in which the model \emph{explicitly} identified
  \emph{both} players’ true strategies (not a “lowest-loss” share). Each match-up has $200$ rounds.}
  \label{tab:strategy_id}
  \vspace{4pt}
  \begin{tabular}{lccc}
    \toprule
    \textbf{Model} & \textbf{Static vs.\ Dynamic} & \textbf{Dynamic vs.\ Dynamic} & \textbf{Human-inspired vs.\ Dynamic} \\
    \midrule
    GPT-4o-mini & 0.0\% & 0.0\% & 0.0\% \\
    o3          & 57.5\% & 41.5\% & 0.5\% \\
    Claude~3.7  & 21.5\% & 0.0\%  & 1.0\% \\
    \bottomrule
  \end{tabular}
\end{table}

\paragraph{Claude 3.7.}
Claude’s explanations consistently report aggregate proportions and align them to \emph{static} templates (e.g., “closest to $M/N$”) while often stating “no clear dynamic pattern.” 
This behavior partially reduces loss in Static--Dynamic (\autoref{fig:union_loss_all}a) with a correspondingly \emph{moderate} SIR in \autoref{tab:strategy_id}, 
but it degrades in Dynamic--Dynamic (\autoref{fig:union_loss_all}b) where sensitivity to opponent-conditioned reactions is required; there SIR collapses to near zero (\autoref{tab:strategy_id}). 
The oscillation between adjacent static labels without engaging the sequential dependency is consistent with its \emph{mid-level, higher-variance} loss curves.

\paragraph{GPT-4o-mini.}
GPT-4o-mini frequently exhibits \emph{description--attribution inconsistency} and unstable commitments, e.g., labeling a distribution as “Paper-biased” while concluding it “matches Rock-biased (H),” and alternating among Rock-biased, Paper-biased, and dynamic $Z$. 
These internally inconsistent or templated narratives rarely culminate in correct dual attributions, producing \emph{near-zero SIR} across settings (\autoref{tab:strategy_id}) and \emph{persistently high} Union Loss with limited round-wise improvement (\autoref{fig:union_loss_all}). 
In D vs.\ Y, it fails to track the one-step, opponent-conditioned countering rule, mirroring the general pattern seen across models.

\begin{table}[ht]
\centering
\caption{Representative reasoning snippets (abridged) produced by each model for the N vs.~G matchup.}
\label{tab:rep_snippets}
\begin{tabularx}{\linewidth}{@{}l c >{\raggedright\arraybackslash}X@{}}
\toprule
\textbf{Model} & \textbf{Round} & \textbf{Reasoning} \\
\midrule
o3 & 40  & P2 0\% R, $\sim$50/50 P+S; matches $G$; no adaptive pattern. \\
GPT-4o-mini & 60 & P1 prefers Paper; this aligns with Rock-biased. P2 mirrors (Z). \\
Claude & 140 & P1 60\% Paper, 28\% Scissors; P2 48\% Scissors, 44\% Paper; no clear dynamic pattern. \\
\bottomrule
\end{tabularx}
\end{table}
Taken together, these findings provide a direct answer to \textbf{RQ2}. 
Our metrics—loss calibration, explicit commitments (SIR), and reasoning consistency—jointly capture competencies that approximate \emph{mind-like inference}. 
They reflect necessary building blocks such as belief updating, probability calibration, and stable strategy identification, which together offer a practical proxy for Theory of Mind without claiming full equivalence. 
Additional reasoning tables and excerpts are provided in App.~\ref{app:reasoning-data} for completeness.

\paragraph{Cross-model comparison.}

Figure~\ref{fig:union_overview} (a–c) compares the \emph{normalized Union loss} of the three models across the three match-ups. 
Two robust regularities emerge. 
First, in both \textbf{H vs.~C} (dynamic rock-biased vs.\ static pure paper) and \textbf{N vs.~G} (dynamic vs.\ dynamic), \textsc{o3} maintains the lowest loss trajectories for most rounds, while \textsc{Claude~3.7} tracks slightly higher and \textsc{GPT-4o-mini} lags behind. 
Second, \textsc{GPT-4o-mini} exhibits an almost flat plateau (roughly $\sim\!0.20$ in normalized Union loss), indicating weak sensitivity to accumulating trajectory evidence and minimal belief updating; by contrast, \textsc{o3} and \textsc{Claude~3.7} show pronounced dips whenever the latent structure becomes identifiable.
This pattern is corroborated by the model-provided explanations in Appendix~\ref{app:reasoning-data}: as shown in Table~\ref{tab:reason-4omini}, \textsc{GPT-4o-mini}'s reasoning remains nearly invariant across rounds, consistent with the flat loss plateau.

\captionsetup[subfigure]{justification=centering,font=small}
\begin{figure}[!tbpt]
    \centering

    \begin{subfigure}[t]{0.32\textwidth}
        \centering
        \vspace{0pt}
        \includegraphics[width=\linewidth]{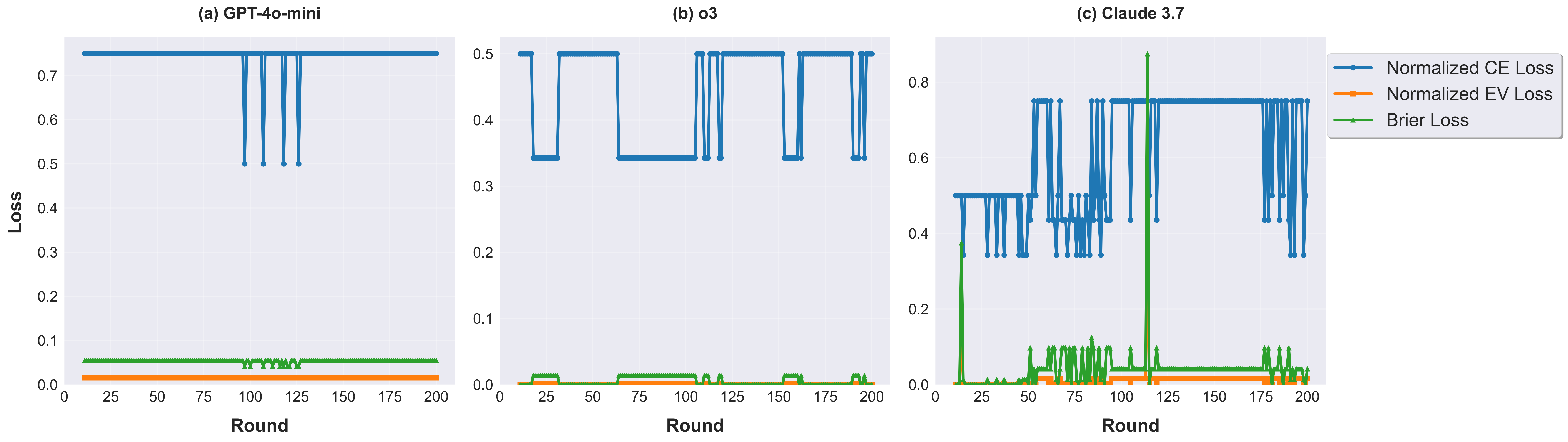}
        \caption{H vs. C setting (dynamic vs. staticr).}
        \label{fig:HC_union_loss}
    \end{subfigure}\hfill
    \begin{subfigure}[t]{0.32\textwidth}
        \centering
        \vspace{0pt}
        \includegraphics[width=\linewidth]{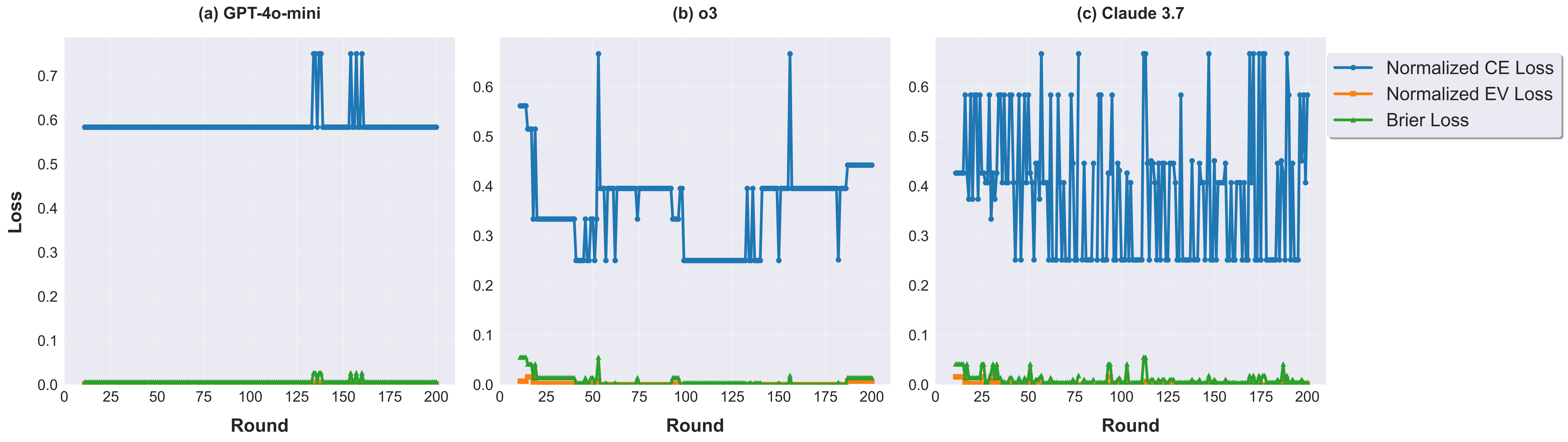}
        \caption{N vs. G setting (dynamic vs. dynamic).}
        \label{fig:NG_union_loss}
    \end{subfigure}\hfill
    \begin{subfigure}[t]{0.32\textwidth}
        \centering
        \vspace{0pt}
        \includegraphics[width=\linewidth]{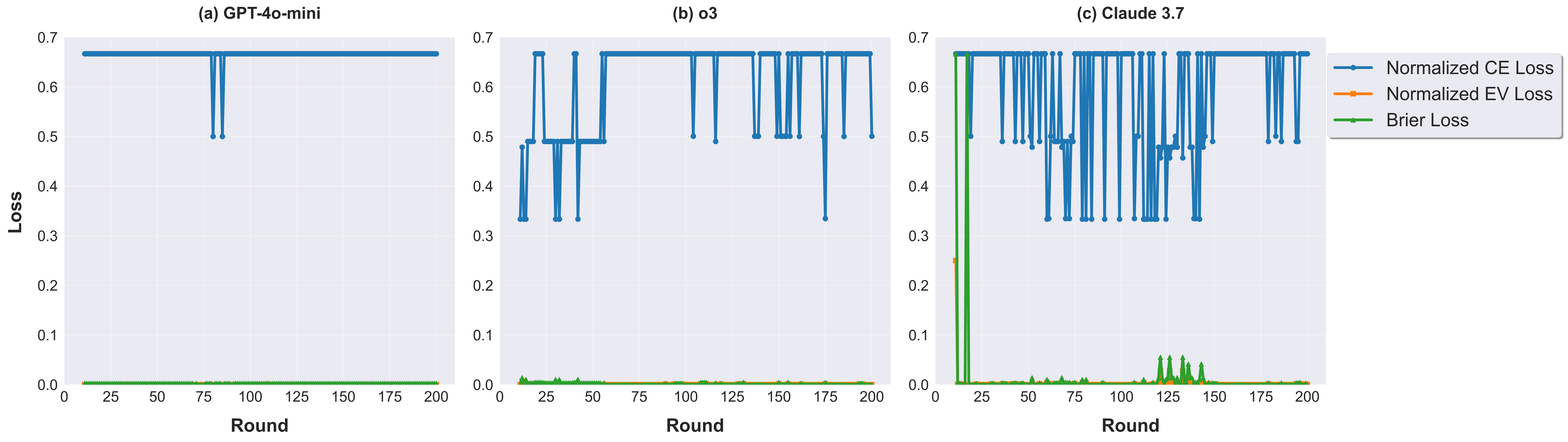}
        \caption{D vs. Y setting (dynamic vs. human-inspired).}
        \label{fig:DY_union_loss}
    \end{subfigure}

    \caption{Comparison of Union Loss across models in different settings.}
    \label{fig:union_loss_all}
\end{figure}

\paragraph{Calibration vs.\ payoff reasoning.}
Looking at the components (CE, Brier, EV Loss~\ref{app:loss-plots}), \textsc{o3} tends to reduce both calibration error (Brier) and payoff discrepancy (EVLoss) in tandem, consistent with its lower Union loss. 
\textsc{Claude~3.7} adapts but exhibits intermittent spikes in CE/EVLoss, consistent with occasional mis-specification before it corrects. 
\textsc{GPT-4o-mini} shows small variance but at a higher plateau across components, consistent with conservative forecasts that under-update rather than exploit deterministic asymmetries (notably in H vs.~C).

\paragraph{Mechanism Identification as an Auxiliary Diagnostic.}
Table~\ref{tab:strategy_id} reports the share of rounds in which each model \emph{explicitly identifies both players’ strategies} (this is \emph{not} “lowest loss per round”). 
The pattern mirrors the loss ranking: \textsc{o3} attains the highest identification rates in \emph{Static vs.~Dynamic} (H vs.~C: $57.5\%$) and \emph{Dynamic vs.~Dynamic} (N vs.~G: $41.5\%$), \textsc{Claude~3.7} is intermediate (e.g., $21.5\%$ in H vs.~C), and \textsc{GPT-4o-mini} is near $0\%$. 
In \emph{Human-inspired vs.~Dynamic} (D vs.~Y), identification drops to $\approx 0.5$–$1.0\%$ even for the stronger models, underscoring that low loss does not necessarily imply that the model can \emph{name} the latent policies.

\begin{table}[t]
  \centering
  \caption{\textbf{Strategy identification rate (\%).}
  Values denote the percentage of rounds in which the model \emph{explicitly} identified
  \emph{both} players’ true strategies in the transcript (not a “lowest-loss” share).
  Each match-up has $200$ rounds.}
  \label{tab:strategy_id}
  \vspace{4pt}
  \begin{tabular}{lccc}
    \toprule
    \textbf{Model} & \textbf{Static vs.\ Dynamic} & \textbf{Dynamic vs.\ Dynamic} & \textbf{Human-inspired vs.\ Dynamic} \\
    \midrule
    GPT-4o-mini & 0.0\% & 0.0\% & 0.0\% \\
    o3          & 57.5\% & 41.5\% & 0.5\% \\
    Claude~3.7  & 21.5\% & 0.0\%  & 1.0\% \\
    \bottomrule
  \end{tabular}
\end{table}

\section{Conclusion and Limitations}
In this work, we introduced an interactive and reproducible demo for probing LLM game understanding through Rock--Paper--Scissors, positioning the model as an \emph{Observer} rather than a player. Our framework evaluates distributional alignment between predictions and ground-truth outcomes using a principled Union Loss that integrates cross-entropy, Brier score, and expected-value discrepancy, thereby capturing both behavioral and utility aspects of reasoning. Real-time visualization and full control over prompts, memory, and candidate strategies make the system transparent, extensible, and suitable as a benchmark for future comparative studies.
While our current evaluation is limited to simple environments such as RPS and focuses primarily on surface-level reasoning traces, the framework offers a foundation for broader exploration. Future work could extend it to more complex, multi-agent, or strategic games, and incorporate deeper diagnostics to assess reasoning chains more systematically. We view this work as an initial step toward building interactive, aiming to create interpretable testbeds for understanding the reasoning capabilities of large language models.

{\small
\bibliographystyle{plainnat}
\bibliography{refs}
}

\clearpage

\appendix

\section{Candidate Pool}
\label{app:player-table}

\begin{table}[!htbp]
\centering
\caption{Rock--Paper--Scissors strategy library.  ``Type'' indicates different types of strategies where S stands for\textbf{S}tatic, D for \textbf{D}ynamic,
and H for \textbf{H}uman (reactive) policy.}
\begin{tabular}{cccccc}
\toprule
\textbf{Key} & \textbf{Name}           & \textbf{Rock} & \textbf{Paper} & \textbf{Scissors} & \textbf{Type}\\
\midrule
A  & Pure Scissors     & 0     & 0     & 1     & S\\
B  & Pure Rock         & 1     & 0     & 0     & S\\
C  & Pure Paper        & 0     & 1     & 0     & S\\
\midrule
D  & Uniform Random    & 0.333 & 0.333 & 0.334 & D\\
E  & Rock + Paper      & 0.50  & 0.50  & 0     & D\\
F  & Rock + Scissors   & 0.50  & 0     & 0.50  & D\\
G  & Paper + Scissors  & 0     & 0.50  & 0.50  & D\\
H  & Rock Biased       & 0.50  & 0.25  & 0.25  & D\\
I  & Paper Biased      & 0.25  & 0.50  & 0.25  & D\\
J  & Scissors Biased   & 0.25  & 0.25  & 0.50  & D\\
K  & Rock $>$ Paper    & 0.50  & 0.333 & 0.167 & D\\
L  & Rock $>$ Scissors & 0.50  & 0.167 & 0.333 & D\\
M  & Paper $>$ Rock    & 0.333 & 0.50  & 0.167 & D\\
N  & Paper $>$ Scissors& 0.167 & 0.50  & 0.333 & D\\
O  & Scissors $>$ Rock & 0.333 & 0.167 & 0.50  & D\\
P  & Scissors $>$ Paper& 0.167 & 0.333 & 0.50  & D\\
\midrule
X  & Win‐Last          & \multicolumn{3}{c}{Reactive policy}  & H\\
Y  & Lose‐Last         & \multicolumn{3}{c}{Reactive policy}  & H\\
Z  & Copy‐Last         & \multicolumn{3}{c}{Reactive policy}  & H\\
\bottomrule
\end{tabular}
\end{table}

\section{Prompt Example}
\label{app:prompt-example}

\begin{lstlisting}
You are an RPS observer. Infer the most likely strategies for P1 and P2 
from the catalog and history. Respond with JSON only.

[Strategy Catalog]
{
  "A": {
    "type": "static", 
    "name": "A (Pure Scissors)", 
    "dist": {"rock": 0, "paper": 0, "scissors": 1}}, 
  .......
  "Z": {
    "type": "dynamic", 
    "name": "Z", 
    "rule": "Play the same move as the opponent's previous move (e.g.,
             if the opponent favored Scissors in the last round, 
             I will also favor Scissors; and so on)."}
}

Notes:
- Static strategies (type=static): fixed move distribution dist={rock,paper,scissors}.
- Dynamic strategies (type=dynamic): depend on opponent's previous move; 
  field 'rule' describes the behavior.

[Game History]
[{"round": 1, "move1": 2, "move2": 1, "result": 1}, 
 {"round": 2, "move1": 0, "move2": 1, "result": -1}, 
 {"round": 3, "move1": 0, "move2": 1, "result": -1}, ...]

Notes: an array, each element contains:
- move1: Player 1 move (0=Rock, 1=Paper, 2=Scissors)
- move2: Player 2 move (0=Rock, 1=Paper, 2=Scissors)
- result: from Player 1 perspective (1=win, 0=draw, -1=loss)

Think step by step:
1. Compute approximate move frequencies for Player 1.
2. Match Player 1's distribution to the closest catalog strategy.
3. Do the same for Player 2.
4. Estimate confidence.
5. Output ONLY the following JSON and nothing else.

{
  "guess_s1": <code like 'H'>,
  "guess_s2": <code like 'Z'>,
  "confidence": <decimal between 0 and 1>,
  "reasoning": <3-5 phrases; separated by semicolons>
}
\end{lstlisting}

\section{Reasoning Data}

\label{app:reasoning-data}
\begin{table}[ht]
\centering
\caption{4o-mini reasoning by round}
\label{tab:reason-4omini}
\footnotesize
\begin{tabularx}{\linewidth}{c X}
\toprule
\textbf{round} & \textbf{reasoning} \\
\midrule
20 & Player 1 has favored Scissors significantly; Player 1's distribution resembles a Rock-biased strategy. Player 2 consistently plays Paper, suggesting a tendency to match Player 1's last move; Player 2's behavior aligns with a strategy that mirrors Player 1's previous move. \\
40 & Player 1 has a higher frequency of Scissors; Player 1's moves suggest a Rock-biased strategy. Player 2 consistently plays Paper, indicating a tendency to match Player 1's last move. The history shows repeated patterns favoring these strategies. \\
60 & Player 1 has a high frequency of Rock; Player 2 consistently plays Paper; Player 1's strategy aligns with Rock-biased strategy H; Player 2's strategy mimics Player 1's last move, indicating Z. \\
80 & Player 1 has a high frequency of Rock; Player 1's strategy resembles Rock-biased (H); Player 2 consistently plays Paper, suggesting a tendency to mirror or counter; Player 2's strategy aligns with the dynamic strategy of mirroring (Z). \\
100 & Player 1 has shown a strong tendency towards Rock; Player 2 consistently plays Paper, indicating a counter-strategy; Player 1's distribution closely matches a random strategy; Player 2's behavior suggests they are mimicking Player 1's last move. \\
120 & Player 1 has a high frequency of Rock and Paper; Player 1's moves suggest a random strategy. Player 2 consistently plays Paper, indicating a tendency to mirror previous moves. The confidence is high due to the clear patterns observed. \\
140 & Player 1 has a high frequency of Rock; Player 1's moves show a tendency towards Rock-biased strategies; Player 2 consistently plays Paper, suggesting a reactive strategy; Player 2's last move was Paper, indicating they may repeat it. \\
160 & Player 1 has a high frequency of Rock; Player 2 consistently plays Paper; Player 1's strategy resembles Rock-biased; Player 2's strategy seems to mimic Player 1's last move. \\
180 & Player 1 has a high frequency of Rock; Player 1's last moves show a tendency towards Rock. Player 2 consistently plays Paper, indicating a reactive strategy. Player 2's last moves suggest they are likely to repeat their last move. \\
200 & Player 1 has a high frequency of Rock and Paper; this aligns with a Rock-biased strategy. Player 2 consistently plays Paper, suggesting a tendency to repeat moves. The history shows Player 1 often counters Player 2's Paper with Rock. \\
\bottomrule
\end{tabularx}
\end{table}
\FloatBarrier

\begin{table}[ht]
\centering
\caption{Claude~3.7~Sonnet reasoning by round}
\label{tab:reason-claude-sonnet}
\footnotesize
\begin{tabularx}{\linewidth}{c X}
\toprule
\textbf{round} & \textbf{reasoning} \\
\midrule
20 & Player 1 uses mostly Rock (13/20 = 65\%) with some Paper (4/20 = 20\%) and Scissors (3/20 = 15\%); closest to H (Rock-biased). Player 2 exclusively plays Paper (20/20 = 100\%) matching strategy C (Pure Paper); distributions are very clear. \\
40 & Player 1 shows Rock bias with approximately 27/40 Rock moves; Player 2 plays Paper exclusively (40/40 moves); static strategies are clear with very consistent patterns; no evidence of dynamic strategy adaptation. \\
60 & Player 1 uses roughly equal mix of Rock, Paper, Scissors; Player 2 plays Paper exclusively in all 50 rounds; P1 distribution matches Random strategy; P2 distribution is exactly Pure Paper. \\
80 & Player 1 used Rock 58\%, Paper 30\%, Scissors 12\% which matches strategy L; Player 2 used Paper 100\% of the time which perfectly matches strategy C; Player 2's consistent behavior makes this a high confidence assessment. \\
100 & Player 1 plays Rock 54\%, Paper 30\%, Scissors 16\%, roughly approximating strategy D (Random); Player 2 plays Paper 100\% of the time, perfectly matching strategy C (Pure Paper); P2's strategy is clearly C with all moves being Paper; P1's distribution is closest to D though slightly biased toward Rock. \\
120 & Player 2 plays Paper (1) in all 50 rounds; Player 1's distribution is approximately 26 Rock (52\%), 12 Paper (24\%), 12 Scissors (24\%), which is close to random with a slight Rock bias; P2 perfectly matches strategy C (Pure Paper); P1 is closest to strategy D (Random). \\
140 & Player 2 plays Paper (1) in all 50 rounds; Player 1's distribution is approximately 26 Rock (0.52), 12 Paper (0.24), 12 Scissors (0.24) which is closest to D (Random); Player 2 is clearly using strategy C (Pure Paper). \\
160 & Player 1 uses a mixed strategy with Rock (18/50), Paper (15/50), Scissors (17/50), approximating a random distribution; Player 2 plays Paper exclusively in all 50 rounds; P2's perfect consistency indicates a pure strategy; P1's distribution closely matches strategy D (Random). \\
180 & P2 plays Paper in 100\% of moves; P1 plays Rock $\sim$60\%, Paper $\sim$25\%, Scissors $\sim$15\%, which is closest to Random with slight Rock bias; P2 perfectly matches Pure Paper strategy; P1 doesn't perfectly match any static strategy but is closest to Random. \\
200 & Player 2 plays Paper (1) in all 50 rounds; Player 1 plays Rock 27 times (54\%), Paper 9 times (18\%), and Scissors 14 times (28\%); P1's distribution is close to random (strategy D); P2 is clearly using Pure Paper strategy (C). \\
\bottomrule
\end{tabularx}
\end{table}

\clearpage
\begin{table}[ht]
\centering
\caption{o3 reasoning by round}
\label{tab:reason-o3}
\footnotesize
\begin{tabularx}{\linewidth}{c X}
\toprule
\textbf{round} & \textbf{reasoning} \\
\midrule
20 & P1 frequencies 0.50/0.15/0.35 align with L template 0.50/0.167/0.333; P1 shows no clear dependence on P2 moves; P2 uses Paper in all 20 rounds matching pure Paper strategy C precisely; sample size 20 supports stable static inferences. \\
40 & P1 plays 21R/9P/10S $\approx$ 0.53--0.23--0.25; matches H's 0.50--0.25--0.25; P2 plays Paper in all 40 rounds, matching pure Paper strategy C; static behavior from both; large sample supports high confidence. \\
60 & P1 frequencies 0.62/0.22/0.16 Rock--Paper--Scissors; smallest distance to H (0.5/0.25/0.25); P2 played Paper 50/50 rounds, exactly strategy C; patterns steady, no dependency on prior moves; thus static match likely; high certainty, especially for P2. \\
80 & P1 frequencies 0.54/0.34/0.12 align with K 0.50/0.333/0.167; P2 shows 100\% Paper consistent with C; no round-to-round dependence seen; static fits outperform dynamic; 50-round sample supports high certainty. \\
100 & P1 moves: 24R/17P/9S $\approx$ 0.48/0.34/0.18; nearest catalog dist is K 0.50/0.333/0.167; no sign of dynamic adaptation; P2 played Paper in 50/50 rounds so matches pure Paper C; constancy rules out dynamic strategies; large sample boosts confidence. \\
120 & P1 plays 44\% Rock, 32\% Paper, 24\% Scissors; nearest catalog match is H 0.50/0.25/0.25; P2 plays Paper in all 50 rounds; matches pure Paper strategy C exactly; both are static patterns; limited sample but clear tendencies. \\
140 & P1 frequencies 0.56/0.22/0.22 close to H 0.50/0.25/0.25; alternative static options farther; no reaction to P2's prior move detected; P2 moves are 100\% Paper matching C; patterns highly consistent, yielding high confidence. \\
160 & P1 frequencies $\sim$62\% R, 24\% P, 14\% S; nearest catalog mix is K (0.50/0.333/0.167); P2 played Paper in all 50 rounds, matching pure Paper strategy C; dynamic rules would show variation, but none observed; sample size gives moderate certainty. \\
180 & Player1 52\% Rock, 26\% Paper, 22\% Scissors $\approx$ H; Player2 used Paper in 50/50 rounds $\rightarrow$ C; no evidence of move-dependent rules; distributions stable over time; strong static match leads to high confidence. \\
200 & P2 played Paper in all 50 rounds, matching pure Paper C; P1 move mix 42\% Rock, 32\% Paper, 26\% Scissors shows Rock bias; this distribution is closest to strategy H (0.50/0.25/0.25); sequences show no reactive or adaptive pattern, indicating static play; therefore H and C are the most likely strategies with high confidence. \\
\bottomrule
\end{tabularx}
\end{table}

\section{Supplementary Component Loss Plots}
\label{app:loss-plots}
\begin{figure*}[t]
  \centering
  \includegraphics[width=\linewidth]{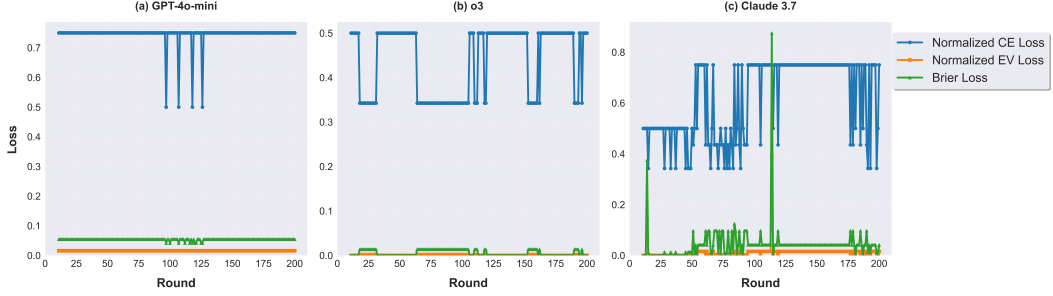}
  \caption{\textbf{H vs.~C:} Normalized component losses over 200 rounds.}
  \label{fig:HC_normloss_row}
\end{figure*}

\begin{figure*}[t]
  \centering
  \includegraphics[width=\linewidth]{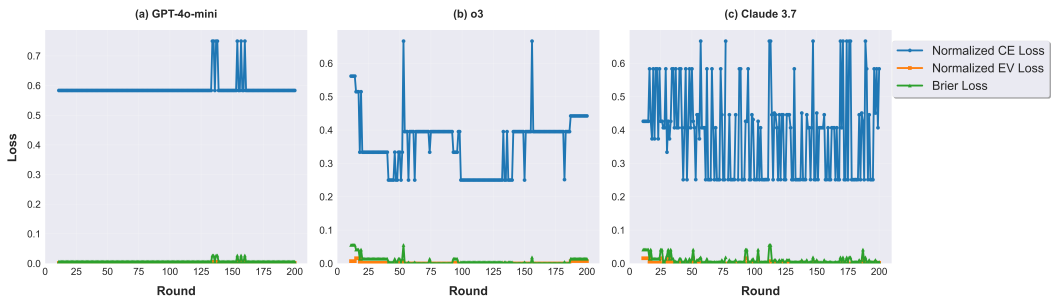}
  \caption{\textbf{N vs.~G:} Normalized component losses over 200 rounds.}
  \label{fig:NG_normloss_row}
\end{figure*}

\begin{figure*}[t]
  \centering
  \includegraphics[width=\linewidth]{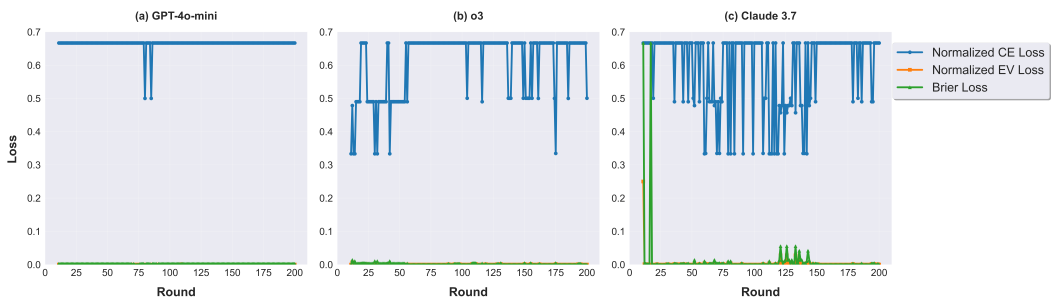}
  \caption{\textbf{D vs.~Y:} Normalized component losses over 200 rounds.}
  \label{fig:DY_normloss_row}
\end{figure*}
\FloatBarrier

\end{document}